\documentclass[letterpaper]{article}

\usepackage{natbib,alifeconf}  

\title{Conditions for Major Transitions in Biological and Cultural Evolution}
\author{Peter D. Turney$^{1}$ \\
\mbox{}\\
$^1$Apperceptual, 485 Rue de Cannes, Suite 4 \\
Gatineau, Quebec, Canada, J8V3X2 \\
peter.turney@apperceptual.com} 

\begin{document}
\maketitle

\begin{abstract}
Evolution by natural selection can be seen an algorithm for generating 
creative solutions to difficult problems. More precisely, evolution by 
natural selection is a {\em class} of algorithms that share a set of 
properties. The question we address here is, what are the conditions 
that define this class of algorithms? There is a standard answer to this 
question: Briefly, the conditions are variation, heredity, and selection. 
We agree that these three conditions are sufficient for a limited 
type of evolution, but they are not sufficient for open-ended evolution. 
By {\em open-ended evolution}, we mean evolution that generates a continuous 
stream of creative solutions, without stagnating. We propose a set of 
conditions for open-ended evolution. The new conditions build on the 
standard conditions by adding fission, fusion, and cooperation. We test the 
proposed conditions by applying them to major transitions in the evolution 
of life and culture. We find that the proposed conditions are able to 
account for the major transitions.
\end{abstract}

\section{Introduction}


In biology, {\em evolution} generally means change in the gene pool of a 
population over time, which can be caused by genetic drift, gene 
flow, natural selection, and other processes. In this paper, when 
we use the term {\em evolution}, we specifically mean {\em evolution by 
natural selection}.

\citet[pp.~5-6]{Brandon1996} states the following three components are 
crucial to evolution by natural selection:

\begin{quote}
\begin{enumerate}
\item Variation: There is (significant) variation in morphological, 
physiological and behavioural traits among members of a species.

\item Heredity: Some traits are heritable so that individuals resemble 
their relations more than they resemble unrelated individuals and, 
in particular, offspring resemble their parents.

\item Differential Fitness: Different variants (or different types of 
organisms) leave different numbers of offspring in immediate or remote 
generations.
\end{enumerate}
\end{quote}

\noindent In the literature, {\em differential fitness} is often called 
{\em selection}. \citet{Godfrey-Smith2007} lists the same three 
components, calling them {\em conditions for evolution by natural 
selection}. 

\citet[p.~409]{Taylor2016} define an {\em open-ended evolutionary system} 
as ``one that is capable of producing a continual stream of novel 
organisms rather than settling on some quasi-stable state beyond which 
nothing fundamentally new occurs,'' where {\em organisms} include ``both 
biological organisms and individuals in artificial evolutionary systems 
in software, hardware, or wetware.'' There is a growing consensus 
\citep{Taylor2016} that the standard three conditions (variation, heredity, 
and selection) are sufficient for a limited type of evolution, but they 
are not sufficient for open-ended evolution (OEE).

\citet{Dennett1995} and \citet{Boden2015} have argued that evolution 
by natural selection is an algorithm for generating creative 
solutions to difficult problems. The promise of this view is that simulated 
evolution can be a source of solutions to human problems, in addition to 
providing us with a better understanding of evolution in nature. The 
problem is that current simulations are not open-ended; they reach 
a point where further processing yields diminishing returns. 

Our goal in this paper is to discover the conditions that enable open-ended 
evolutionary systems. \citet[pp.~415-416]{Taylor2016} provide a comprehensive 
list of the {\em behavioural hallmarks} of OEE. To evaluate our proposed 
conditions, we use their hallmark 1(c), {\em major transitions in evolution}. 
The tests for our proposed conditions are seven major transitions in 
biological evolution from \citet{MaynardSmith1995}, seven major 
transitions in human cultural evolution from \citet{Nolan2010},
and one super transition (the evolution of human language) that bridges
biological and cultural evolution \citep{Deacon1998,Richerson2005}.

We selected the major transitions in biological and cultural evolution as 
our hallmark for several reasons. First, the very idea of OEE comes from 
observing evolution in biology and culture (that is, evolution {\em in vivo}, 
as opposed to {\em in vitro} or {\em in silico}); therefore it seems plausible 
that we can learn what OEE requires by studying major transitions in biology 
and culture. Second, the major transitions are relatively clear, whereas 
other hallmarks involve concepts, such as {\em novelty} and {\em complexity}, 
that lack consensus definitions and measures. Third, focusing on 
abstract measures of complexity and novelty may yield conditions that have 
theoretical interest but are irrelevant for understanding biological and
cultural ({\em in vivo}) evolution. 

In the next section, we will present our proposed conditions for major
transitions. The following section will examine related work. The next
three sections will test the conditions by applying them to seven major 
transitions in biological evolution \citep{MaynardSmith1995}, one 
super transition in biocultural coevolution \citep{Deacon1998,Richerson2005}, 
and seven major transitions in human cultural evolution \citep{Nolan2010}. 
We will then summarize our results, discuss future work, and conclude.

\section{Conditions for Major Transitions}

Before we present our conditions for major transitions, we need to 
discuss the kinds of conditions that we are seeking. First, for the sake
of simplicity, we are seeking conditions that make the transitions
{\em possible}, not the conditions that make them {\em probable}.
We assume that the conditions for possible open-endedness are a subset
of the conditions for likely (highly probable) open-endedness. For example,
a given transition might be likely given a very large population, but
merely possible given a population of arbitrary size. Second, we
are seeking {\em declarative} conditions (general, abstract requirements),
not {\em procedural} conditions (step-by-step recipes), because we are
interested in a general class of algorithms, not a specific algorithm.
We assume that there are many different systems that can manifest OEE,
and we wish to characterize the general class, not a special case.

\citet{Simon1962}, in his well-known parable of two watchmakers, 
argued that we should expect a complex system to be composed of relatively 
stable intermediate components, because a system without such components 
would be fragile and unlikely to evolve. Living organisms, human-built 
machines, social organizations, and human languages are composed of 
stable parts, forming a connected whole. In complex systems, parts can 
be decomposed into sub-parts, forming a nested hierarchy. 

\citet{Koestler1967} coined the word {\em holon} for a thing that can 
be seen as either a whole or a part, depending on the level in the 
hierarchy that is the focus of our attention.
In the following description of the conditions for major transitions, 
our basic unit is the holon. Examples of holons in biology 
are cells, animals, plants, organs (parts in plants and animals), and 
organelles (parts in cells). Examples of holons in culture are tools 
(languages, machines), social organizations (families, governments, 
companies), individuals (parts in societies), and components (parts 
in machines).

\begin{enumerate}
\item Reproduction: There are two types of reproduction, which we may
think of as asexual and sexual, although we generalize these concepts
so that they apply to both biology and culture. Reproduction does not
change the number of levels in the part-whole hierarchy of a holon.
(A) A new holon is created from a single parent holon (as in asexual 
reproduction in biology). The child holon has the same parts and 
sub-parts as the parent (typically with some heritable variation; see~5). 
(B) A new holon is created from multiple, similar parent holons (as in 
sexual reproduction in biology, but cultural reproduction may involve 
more than two parents). Each part in the child holon has traits that 
are a blend of the traits of the corresponding parts in any or all of 
the parents (typically with some heritable variation; see~5).

\item Fission-Fusion: Unlike reproduction, fission and fusion may
change the number of levels in the part-whole hierarchy of a holon.
(Fission or fusion may occur during reproduction, but we think of
them as distinct from reproduction.)
(A) With fission, a holon divides into component parts. Depending
on the details of the fission process, the resulting holons may
have fewer levels in their part-whole hierarchy than the original
holon, which no longer exists after fission.
(B) With fusion, two or more holons combine to form a new holon, in 
which the original holons become parts (as in symbiosis in biology). 
Depending on the details of the fusion process, the resulting holon 
may have more levels in its part-whole hierarchy than the original
holons, which do not exist independently after fusion.

\item Differential Fitness: Heritable variation in holons results in 
different numbers of offspring in immediate or remote generations.

\item Cooperation: Open-ended evolution requires the ongoing emergence 
of mechanisms that cause parts of a holon to sacrifice (some of) their 
own differential fitness to support the differential fitness of the whole.
For example, when fusion combines holons, the fusion will generally
require a mechanism that enforces cooperation among the new parts;
otherwise the new holon will have relatively low differential fitness.

\item Heritable Variation: Variation that can be passed on to future 
generations includes (A) change to the traits of holons 
or to the traits of parts of holons, (B) deletion of parts in a holon, 
(C) duplication of parts in holons, (D) changes in reproductive mechanisms 
(see 1), (E) changes in fission or fusion mechanisms (see~2), and 
(F) changes in the mechanisms by which cooperation is enforced (see~4). 
(G) When a new holon is formed by fusion, the fusion of these specific parts 
can be a heritable variation (see~2). (H) When a new holon is formed by 
fission, the existence of this holon as a separate individual can be a 
heritable variation (see~2). 
\end{enumerate}

Comparing Brandon's three conditions for evolution by natural selection
\citep[][see above]{Brandon1996} with our five conditions for open-ended 
evolution (OEE), we see that Brandon's first condition implies sexual 
reproduction, since it mentions {\em members of a species}. Thus his 
conditions relate to ours as follows:

\begin{enumerate}
\item Brandon's Variation $\rightarrow$ Turney's Reproduction (1B) 
and Heritable Variation (5A, 5B, and 5C)

\item Brandon's Heredity $\rightarrow$ Turney's Reproduction (1B) 
and Heritable Variation (5A, 5B, and 5C)

\item Brandon's Differential Fitness $\rightarrow$ Turney's 
Differential Fitness (3)
\end{enumerate}

\noindent Brandon's conditions omit asexual reproduction (1A), fission (2A),
fusion (2B), cooperation (4), and some types of heritable variation (5D,
5E, 5F, 5G, and 5H).

\citet{Brandon1996} assumes a fixed, two-level, part-whole hierarchy, 
consisting of {\em individuals} that are members of a {\em species}. Therefore his 
conditions cannot account for an evolutionary transition that involves a change 
in levels, such as the transition from single-celled creatures
to multi-celled organisms. We shall see that several of the major 
transitions require fusion and cooperation, which enable a shift in the
selection unit (a shift in the level of selection).

\section{Related Work}

Related work falls into four categories: various proposals for conditions
for evolution, arguments for the importance of part-whole hierarchies
in understanding evolution, the role of fission and fusion in
evolution, and work on the evolution of cooperation.

{\bf Conditions for evolution:} \citet{Godfrey-Smith2007,Godfrey-Smith2011}
surveys several different sets of conditions for biological evolution and 
discusses their problems. His own suggested conditions are similar to those 
of \citet{Brandon1996}. \citet{Sterelny2011} proposes eight conditions 
for biological evolution. It seems to us that his conditions lack 
generality, due to their focus on specific biological mechanisms. 

\citet{Taylor2015} and \citet{Soros2014} present conditions for 
open-ended evolution in natural and artificial systems. \citet{Nolan2010} 
present seven conditions for innovation in human societies. None of these 
conditions discuss the necessity of part-whole hierarchies and cooperation.
\citet{MaynardSmith1995} discuss part-whole hierarchies and cooperation,
but they do not attempt to formulate the conditions required for major
transitions.

{\bf Part-whole hierarchies:} \citet{Simon1962} argued for part-whole 
hierarchies in complex systems in the social, biological, and physical 
sciences. \citet{Turney1989} formalized the argument with graph theory.
\citet{McShea2011} assert that the increase in complexity of 
organisms over time is largely due to heritable variation in part-whole 
hierarchies. \citet{Banzhaf2016} define an architecture for building 
artificial life simulations, in which part-whole hierarchies play a 
central role.

{\bf Fission and fusion:} Fusion (2B), forming a new holon by combining 
two existing holons, enables part-whole hierarchies to add new levels 
to the hierarchy. Symbiosis, a kind of fusion, is a core element of 
the major transitions in biology \citep{MaynardSmith1995}. 
\citet{Margulis1970,Margulis1981} played a major role in recognizing
the importance of symbiosis. Fission and fusion (2AB) are also major 
features of cultural evolution \citep{Nolan2010}.

{\bf Cooperation:} It is widely recognized that cooperation plays
an important role in both biological and cultural evolution 
\citep{Axelrod1981,Hammerstein2003}. \citet{MaynardSmith1995}
spend much effort on explaining the various mechanisms by which
cooperation is enforced in biological organisms. The parts in a holon
often have conflicting interests, which require organisms to
evolve ways to subordinate the differential fitnesses of the parts to
the differential fitness of the whole.

\section{Major Transitions in Biological Evolution}

In their highly influential work, \citet[pp.~3-14]{MaynardSmith1995} argue 
that the increase in complexity of biological organisms over time is mostly 
due to a small number of major transitions in evolution. They state 
that the theme unifying these transitions is changes in the way genetic 
information is passed on from one generation to the next. We present 
their major transitions and discuss how our five conditions apply 
to them.

\citet{Szathmary2015} later revised the list of transitions, dropping
sex from the list, because it did not fit his model of major transitions.
We use the original list, including sex, as it seems to us that 
the more transitions our proposed conditions can handle, the better.

{\bf Molecular compartments:} In the right chemical environment, some 
molecules can make copies of themselves by a sequence of chemical 
reactions with other molecules; such molecules are said to be 
{\em autocatalytic}, since they catalyze their own production. A slight 
change to an autocatalytic molecule could increase the efficiency of its 
chemical reaction, which would result in a kind of evolution, since we 
would have variation, heredity, and differential fitness in favour of 
the more efficient reaction. However, this change would be a 
limited form of evolution, because only a small number of variants will 
be able to support the autocatalytic reaction; almost all changes would 
end the reaction. \citet{MaynardSmith1995} call this {\em limited heredity}, 
since only a small number of variations are heritable. 

A way around this limitation is a {\em hypercycle}, consisting of a cycle of 
self-replicating molecules, each of which catalyzes the creation of the 
next molecule in the cycle. Although each molecule has limited heredity, 
there is the possibility of adding more molecules to the hypercycle. 

A difficulty with hypercycles is that they are vulnerable to parasitic 
replicators. {\em Molecular compartments}, by limiting the number of molecules 
within a compartment, can penalize parasitic replicators in a hypercycle 
and force cooperation. \citet{MaynardSmith1995} argue that molecular 
compartments are the key development that allowed the transition from 
limited heredity replicators to {\em unlimited heredity} replicators. 

An autocatalytic molecule is capable of a kind of asexual reproduction
(see~1A), although it has limited heritable variation (5A).
A hypercycle is a fusion of separate autocatalytic reactions (2B and 5G). 
Molecular compartments are a mechanism for enforcing cooperation among
the autocatalytic molecules (4 and 5F). If an artificial evolutionary
system can model autocatalytic molecules and it satisfies our five conditions,
then it seems possible for hypercycles to evolve from autocatalytic molecules.
The five conditions allow this major transition.

{\bf Chromosomes:} In modern cells, genes are linked in chromosomes, 
but it is believed that genes were not linked in the earliest 
protocells. From the {\em selfish gene} perspective \citep{Dawkins1976}, 
it is necessary to explain why one gene would bind its fate to that of 
another gene. It takes longer to replicate two linked genes than to 
replicate either of the genes separately, which puts the linked genes at a 
disadvantage. \citet{MaynardSmith1995} contend that two linked genes 
can out-compete the separate genes when both genes are required for 
efficient reproduction of the protocell and the number of molecules in 
the protocell is small. 

Linking genes is a kind of fusion (2B and 5G). Limiting the number of 
molecules in the protocell is a mechanism for cooperation (4 and 5F). 
Thus chromosomes can evolve in a system that satisfies the five conditions.

{\bf DNA and protein:} All living cells today use DNA as a replicator 
and proteins as enzymes. It is generally accepted that early life was 
based on RNA, which can function as both a replicator and an enzyme 
for catalyzing chemical reactions. This is called the {\em RNA world} 
hypothesis. 

One theory is that modern cells evolved from RNA world in two steps: 
First, cells with RNA alone evolved into cells with RNA and protein; 
later, cells evolved with RNA, protein, and DNA \citep{Forterre2005}. 
In the first step, a cell with an RNA genome would benefit from the ability 
to generate protein enzymes, which are more efficient than RNA enzymes. 
In the second step, switching to a DNA genome would enable cells to 
generate a greater variety of proteins. The first step would facilitate 
the switch, because DNA can be created from the action of protein 
enzymes on RNA.  

A challenge with this theory is to explain how a DNA genome could 
replace an RNA genome. One possibility is that a cell with an RNA 
genome was infected by a DNA virus. Over many generations, the viral 
DNA genome took over the functions of the RNA genome \citep{Forterre2005}. 

When a cell with an RNA genome was infected by a DNA virus, the result 
was a kind of fusion (2B and 5G). Initially cooperation was enforced 
because the DNA virus relied on the host cell for reproduction (4). 
Eventually the RNA genome was eliminated as it became redundant (5B).

{\bf Eukaryotes:} {\em Prokaryotes} are single-celled organisms, including 
bacteria and archaea. {\em Eukaryotes} may be single-celled or multi-celled 
organisms, including protists, fungi, plants, and animals. Eukaryotic 
cells have more complex internal structures than prokaryotic cells, 
including various {\em organelles} (little organs) that are wrapped in 
membranes. 

Every eukaryotic cell has a {\em nucleus}, which is the organelle that 
contains the main genetic material of the cell. Most eukaryotic cells 
contain many {\em mitochondria}, organelles that provide energy to the cell. 
Many plant cells contain {\em chloroplasts}, organelles that perform 
photosynthesis. Mitochondria and chloroplasts were once independent 
prokaryotes that were taken inside host prokaryotes to eventually become 
organelles \citep{MaynardSmith1995}. 

The merging of prokaryotes to form eukaryotes is a kind of fusion (2B and 
5G), called {\em endosymbiosis}. The fusion is mutually beneficial (4): 
The mitochondria and chloroplasts provide energy to their hosts and the 
hosts provide key proteins in return. Some of the genes of mitochondria 
and chloroplasts have migrated into their host's genome, which enforces 
the fusion (4 and 5F).

{\bf Sex:} Asexual cloning has many advantages over sexual reproduction: 
If an organism is better adapted to its environment than 
other organisms of its kind, then its clone will be equally well 
adapted, whereas its child by sexual reproduction is likely to 
be less well adapted. Sexual reproduction brings with it the risks 
of sexually transmitted diseases. Finding a sexual partner takes 
time and effort that can be avoided by parthenogenesis. Sexual 
reproduction leads to wasteful displays, such as the peacock’s tail. 
In plants, sexual reproduction leads to reliance on insect pollinators. 
 
Of the various theories about sex, the evidence appears to support the
hypothesis that sexual reproduction allows a beneficial new mutation to 
spread in a population while also maintaining the variation of genes in 
the population \citep{Keightley2006}. In a population with asexual
reproduction, a beneficial mutation might not be favoured by selection
if it happens to be combined with a harmful mutation. Sexual recombination
can bring together beneficial mutations and split apart harmful mutations.

Our conditions allow sex as an option (1B) without specifying the mechanism.
Whether an organism is sexual or asexual is a heritable variation (5D)
that can change to adapt to changes in the environment.

{\bf Multicellularity:} Multicellularity allows organisms with greater 
complexity and adaptability. Three kinds of eukaryotes have 
independently evolved multicellularity: animals, plants, and fungi.
An animal has many types of cells with varied functions and 
specializations, such as blood cells, nerve cells, and muscle cells. 
\citet{MaynardSmith1995} argue that multicellularity required three 
key developments: gene regulation, cell heredity, and the evolution of 
form. The first two of these three developments are present in 
prokaryotes, but only eukaryotes are multi-celled, so it seems that 
the evolution of form was a crucial step. 

The {\em form} of a multi-celled organism is the spatial distribution of the 
various specialized cells: the shape and structure of the organism. 
Form is controlled by releasing various chemical signals during the 
development of an embryo. A chemical signal is released at a specific 
point and diffuses outwards from that point, resulting in a 
concentration gradient. The local concentration around a cell 
determines which genes will be active in that cell. A carefully timed 
sequence of chemical releases determines how the embryo develops. The 
initial chemical releases determine the general form of the organism 
and later releases determine the details. 

Multicellularity is an instance of fusion (2B and 5G) in which chemical
gradients control the growth and development of the organism (4 and 5F).
Cancerous cells escape the control systems of an organism and reproduce
at the cost of the health of the organism, although the differential
fitness (the number of offspring) of a cancerous cell increases in the 
short term (until the organism dies). The immune system helps to eliminate 
cancerous cells \citep{Corthay2014}, further enforcing cooperation 
(4 and 5F).

{\bf Eusociality:} Three characteristics define eusocial animal 
societies \citep{MaynardSmith1995}: (1) There is a division of labour 
into reproductive and non-reproductive castes. (2) There are 
overlapping generations within a mature colony. (3) There is 
co-operative care for the young; some individuals care for the 
offspring of other individuals. Organisms that have these 
characteristics include insects (ants, bees, wasps, and termites), 
mammals (naked mole-rats), and crustaceans (snapping shrimp). 

From the selfish gene perspective \citep{Dawkins1976}, the puzzle 
is how non-reproductive castes could evolve. This is explained by the
concept of {\em inclusive fitness} \citep{Hamilton1964}. An 
individual's child carries half of the individual's genes, whereas 
the individual's niece or nephew carries one quarter of the individual's 
genes. Therefore, from the selfish gene perspective, two nieces or 
nephews have the same value as a single child of one's own. In a colony 
with one queen, where all individuals are closely related, an individual 
can spread more copies of its genes by caring for the young of others 
than it could by raising children of its own. 

Another puzzle is that there are some colonies with as many as 100 
queens per nest, where individuals might be only distantly related. 
In multi-queen colonies, it turns out that the workers are sterile, 
whereas workers in single-queen colonies are typically fertile.

Eusocial societies are a kind of fusion (2B and 5G). In colonies with
one queen, the mechanism for cooperation is a set of cooperative
behaviours. With multiple queens, cooperation is reinforced
by making workers sterile (4 and 5F).

\section{A Super Transition in Biocultural Coevolution}

Human language may be seen as a {\em super transition}, since language
is the evolutionary link between biology and culture
\citep{Deacon1998,Richerson2005}. Cultural inheritance can take place without 
language; for example, young chimps learn to use sticks to pull ants out of 
their nests by imitating adults. However, imitation only allows {\em limited 
heredity}, compared to the {\em unlimited heredity} of language. 
\citet{MaynardSmith1995} note the analogy between the genetic code 
and human language: They both enable unlimited heredity by using a 
linear sequence of a small set of discrete units. 

Given that language enabled the transition from biological evolution 
to cultural evolution, it is natural to ask, how much of language is 
determined by biology and how much is determined by culture? The 
answer is that the biological and cultural aspects of language 
co-evolved; hence they cannot be cleanly divided. Our vocabulary is 
learned, but certain aspects of our grammar seem to be genetically 
determined. 

The role of genes in grammar raises the question, if a mutation gives 
an individual a new grammatical ability, what use is that new ability 
given that nobody else in the population has it? \citet{MaynardSmith1995} 
address this problem by appealing to the {\em Baldwin effect} 
\citep{Hinton1987}, which explains how behaviours that are originally 
learned can become innate, a process known as {\em genetic assimilation}. 
A new grammatical ability is at first learned, with some cost in terms 
of time and effort spent learning. If a new mutation reduces 
or eliminates the learning effort, that new mutation can spread through 
the population, and the learned grammatical skill can become innate.

Chimps and humans both form social groups, but language allows greater
cooperation among the members of human social groups: Language is a 
mechanism for strengthening cooperation in a social group (4). It 
enhances the fusion of the group (2B), enabling more complex group 
behaviours. The human brain has evolved a variety of mechanisms for 
supporting language (5F). Some aspects of language are inherited by 
genetic mechanisms and other aspects are inherited socioculturally (5G).

The evolution of language appears to have been related to the shift to an 
omnivorous diet, especially big-game hunting. Big-game hunting requires 
planning, cooperation, and communication, all of which benefit from 
improvements in language. Chimpanzees in the wild communicate most 
frequently during meat distribution, which suggests that negotiation of 
meat distribution may have contributed significantly to the evolution of 
language in hunter-gatherer societies \citep{Nolan2010}. 

Language is the primary technology that characterizes hunter-gatherer 
societies. Language provided a mechanism for planning, cooperation, and 
communication that enabled a more effective and efficient fusion of
individuals into a society (2B and 4). New social structures in 
hunter-gatherer societies \citep{Nolan2010} provided an environment that 
encouraged the cultural inheritance of language and knowledge transmitted 
by language (5F and 5G).

\section{Major Transitions in Cultural Evolution}

Human cultural evolution can be viewed in terms of the evolution of  
social organizations (such as families, governments, and corporations) 
or in terms of the evolution of human creations (such as technologies, 
sciences, arts, and languages). \citet{Nolan2010} integrate these two 
views of culture in a principled way. They argue that each major type 
of social organization is characterized by its {\em subsistence 
technology}; that is, the primary technology by which it maintains its 
way of life. For example, simple horticultural societies are characterized 
by the domestication of plants. 

There are other well-known views of human history, such as the work of
\citet{Diamond1997} and \citet{Harari2015}, but they do not attempt to 
systematically divide cultural developments into periods that could be
described as major transitions in cultural evolution. We believe that
\citet{Nolan2010} best define the major transitions in 
human cultural evolution; therefore the following transitions are 
based on their work. We present seven major transitions in cultural 
evolution and apply our five conditons to them.

{\bf Domestication of plants:} {\em Simple horticultural societies} 
emerged around 10,000 to 8,000 BC. Horticultural societies clear land 
by cutting and burning wild trees and shrubs. The ash provides fertilizer 
for the crops that they plant. After a few seasons of farming, the ash 
is depleted and the land must be abandoned until the trees and shrubs 
grow back. 

People in horticultural societies work harder and have less 
freedom than people in hunter-gatherer societies, so it is not obvious 
why people would switch from hunting and gathering to horticulture. 
\citet{Nolan2010} argue that three factors led to the transition: 
(1) environmental change and excessive hunting altered the distribution 
and reduced the population of large game animals, (2) human population 
growth resulted in increased demand for food and horticulture was better 
able to meet the demand than hunting and gathering, and (3) improvements 
in domestication made horticulture increasingly effective. 

Horticulture increased the permanence of human settlements, allowing 
people to accumulate more tools, weapons, and material goods. Horticulture 
allowed larger settlements and denser populations, which resulted in 
growth in trade and commerce. A reliable surplus of food and larger 
populations allowed specialized occupations (e.g., butcher, bead maker, 
tool maker) and increased the rate of innovation.

Domestication of plants allowed larger societies (2B), which in turn
allowed specialized occupations, analogous to organs in a multi-celled
organism. A specialist can acquire a degree of expertise that is not 
possible for a generalist, which increases the differential fitness
of the society. Specialization also enforces cooperation (4), because
a specialist must rely on the skills of other specialists, whereas a
generalist is more self-reliant. From a biological perspective, 
horticulture is a symbiotic fusion of plants and humans (2B), based 
on the technology of stone tools for cutting and fire for burning 
wild trees and shrubs (4).

{\bf Domestication of animals:} {\em Herding societies} emerged around 
the same time as simple horticultural societies \citep{Nolan2010}. Many 
societies practiced both horticulture and herding, but, in some areas, 
crop cultivation was limited due to lack of rain, short growing seasons, 
or mountainous land. Herding usually involves a nomadic lifestyle, as the 
herd exhausts the local pasture. Herding people began riding camels or 
horses around 2000 BC, which allowed relatively large societies to form 
in the open grasslands. Horses, camels, and human population size gave 
herders a military advantage over other societies of the time.

Domestication of animals allowed larger societies (2B), in comparison 
with the hunter-gatherer societies that they replaced. Larger societies 
enabled specialization, which enforces cooperation (4). From a biological 
perspective, herding is a symbiotic fusion of animals and humans (2B), 
based on the skills and tools of shepherding (4).

{\bf Nonferrous metallurgy:} {\em Advanced horticultural societies} are 
characterized by their use of nonferrous metals for the manufacture of 
weapons and tools \citep{Nolan2010}. Around 7000 BC, copper nuggets were 
hammered, without heating, into small tools and ornaments. The earliest 
advanced horticultural societies appeared around 4000 BC, when annealing 
copper with heat allowed the creation of less brittle tools for a wide 
range of purposes. 

Further developments were smelting copper ore, casting, 
and making bronze (alloys of copper with tin or other metals, harder and 
more durable than pure copper). Bronze weapons made warfare highly 
profitable and resulted in large, stratified societies, with an 
aristocratic ruling class, a common class, and often a large number of 
captive slaves, taken in wars. Large societies (2B) allowed specialized 
occupations and required an administrative class (4), which resulted in 
further innovations, such as writing, money, and irrigation.

{\bf Plowing:} {\em Simple agrarian societies} use plows to turn over 
soil, instead of the hoes used by horticultural societies (Latin 
{\em horti cultura}: the cultivation of a garden; {\em agri cultura}: 
the cultivation of a field) \citep{Nolan2010}. Plows reach a greater 
depth than hoes, which brings nutrients to the surface where plants 
can reach them, and also buries weeds, converting them to humus and 
aiding plant growth. Hoed gardens must be abandoned after a few years, 
when the nutrients are depleted, whereas plowed fields last longer. 

The earliest plows appeared around 3000 BC. At first, plows were pulled 
by people, but eventually oxen were used, making it possible for a 
farmer to cultivate a much larger area. The increased productivity 
resulted in a large economic surplus, enabling more complex forms of 
social organization (2B and 4).

{\bf Iron metallurgy:} {\em Advanced agrarian societies} use iron for tools 
and weapons \citep{Nolan2010}. Iron is stronger than bronze and the scarcity 
of tin limits the supply of bronze, but the technology for smelting iron 
ore is relatively complex and it did not become common until about 1200 BC. 
Later it was discovered that iron could be hardened by adding carbon 
and by quenching the hot metal in water. With these developments, iron 
replaced bronze as the preferred material for tools and weapons. This 
accelerated the trend to larger and more stratified societies (2B and 4).

{\bf Fossil fuel energy:} {\em Industrial societies} derive the majority 
of their income from goods produced with fossil fuel energy. Around 1800 
AD, England became the first industrial society when industry powered by 
coal became more economically important than agriculture \citep{Nolan2010}. 
The dominant industries in England at that time were textiles and iron 
making. Beginning around 1760, the textile industry became increasingly 
mechanized. Human-powered machines were replaced by larger, faster, more 
complex machines, powered by steam engines burning coal. 

Between 1770 and 1845, the contribution of the textile industry to 
the national economy increased by a factor of five. The iron 
industry switched from wood to coal for smelting and refining iron ore. 
Iron production in England went from 68 thousand tons in 1788 to 1.6 
million tons in 1845.

In England before 1760, spinning and weaving were cottage industries. 
A family with a spinning wheel and a loom could make cloth in their 
own home. With the introduction of steam engines (burning coal for 
power) made with iron (smelted and refined with coal), large factories 
could make cloth much more efficiently, but the manufacturing process 
became significantly more complex, involving at least three new 
industries (coal mining, steam engine manufacturing, and iron mining 
and processing) and a much larger scale of organization (2B and 4). 

{\bf Information and communication technology:} {\em Information 
societies} generate more wealth from the service sector of the 
economy than the manufacturing sector; knowledge, information, and 
communication surpass energy in importance \citep{Nolan2010}. In 
the period from 1986 to 2007, world computation capacity (in MIPS 
per capita) increased 58\% annually, telecommunications capacity 
(in MB per capita per day) increased 28\% annually, and storage 
capacity (in MB per capita) increased 23\% annually \citep{Hilbert2011}. 
Population growth over the same period was 1\% to 1.5\% per year and 
economic growth was about 6\% to 8.5\% per year \citep{Hilbert2012}. 

\citet{Hilbert2012} estimates that 2002 was the year the 
worldwide quantity of stored digital information exceeded the 
quantity of stored analog information. In 2007, 97\% of all stored 
information was digital \citep{Hilbert2012}. Transforming information 
to digital makes it more accessible and useful by enabling 
computers to search, index, transmit, and analyze the information. 
With computation, stored information becomes a dynamic resource 
instead of a static record.

If we view cultural transitions from the perspective of the organization 
of human societies, each transition has resulted in a significant 
increase in population size and societal complexity (2B). The number of 
specialized crafts and occupations also increases with each transition 
(4). Each transition adds new levels of social organization. If we view 
cultural transitions from the perspective of technology, we also see 
increasing complexity and new levels of organization. With written
and spoken language and visual representations (architectural drawings,
circuit diagrams, chemical formulae, mathematical formulae, and so on), 
these changes are heritable and open to variations (5A-H).

\section{Discussion}

From our analysis of major transitions in biological and cultural 
evolution, it seems that our five conditions are sufficient 
to account for the transitions. On the other hand, Brandon's three 
conditions seem insufficient \citep[][see above]{Brandon1996}. 
In our analysis of the major transitions, most of them 
involve fusion (2B and 5G) and cooperation (4 and 5F), which are 
not covered by Brandon.

A core argument of \citet{MaynardSmith1995} is that much of evolutionary 
theory has focused on the evolution of plants and animals with sexual 
reproduction; before their work, evolutionary theory tended to ignore 
major transitions. Therefore it is not surprising that \citet{Brandon1996}
does not account for the major transitions of \citet{MaynardSmith1995}.
The recent surge of interest in open-ended evolution \citep{Taylor2016}
has brought more attention to the need for understanding the
conditions for major transitions.

In the transitions above, fusion and cooperation play key roles, but
fission does not play a large role in the transitions we have discussed here.
However, fission plays an important role in cultural evolution, such as in the 
human migrations out of Africa, to Oceania, Europe, Asia, the Americas, the 
Pacific, and the Arctic. Fission also plays an important role in the evolution 
of human social organizations, such as the formation of new religious 
organizations from existing organizations and the formation of new companies 
from subsidiaries of existing companies.

\section{Future Work}

A natural next step in this research would be to develop a software
simulation that satisfies our five conditions and then run various
experiments to see how the simulation behaves. As far as we know,
there is no existing simulation that satisfies all of our conditions.

One type of experiment would be to simulate a specific major transition
by initializing the simulation with a simplified model of the situation
before the transition, and then see whether the simulation is able to 
evolve into a simplified model of the situation after the transition.

\section{Conclusion}

Our goal was to find conditions that define a class of algorithms
capable of open-ended evolution. As our behavioural hallmark of OEE
\citep{Taylor2016}, we chose major transitions in evolution. Our analyses 
of major transitions in biological evolution, cultural evolution, 
and biocultural coevolution suggest that the five conditions presented here 
are sufficient to account for the transitions. We expect that these five 
conditions (viewed as cultural products) will themselves evolve over time, 
just as these conditions have evolved from the work of other researchers 
\citep{Axelrod1981,Brandon1996,MaynardSmith1995}.

\section{Acknowledgements}

Thanks to Martin Brooks, Daniel Lemire, Tim Taylor, Shaun Turney, and 
Andr\'e Vellino for very helpful comments on earlier versions of this paper.

\footnotesize
\bibliographystyle{apalike}
\bibliography{turney-oee3} 

\begin{thebibliography}{}

\bibitem[Axelrod and Hamilton, 1981]{Axelrod1981}
Axelrod, R. and Hamilton, W.~D. (1981).
\newblock The evolution of cooperation.
\newblock {\em Science}, 211:1390--1396.

\bibitem[Banzhaf et~al., 2016]{Banzhaf2016}
Banzhaf, W., Baumgaertner, B., Beslon, G., Doursat, R., et~al. (2016).
\newblock Defining and simulating open-ended novelty: Requirements, guidelines,
  and challenges.
\newblock {\em Theory in Biosciences}, 135:131--161.

\bibitem[Boden, 2015]{Boden2015}
Boden, M.~A. (2015).
\newblock Creativity and {ALife}.
\newblock {\em Artificial Life}, 21:354--365.

\bibitem[Brandon, 1996]{Brandon1996}
Brandon, R.~N. (1996).
\newblock {\em Concepts and methods in evolutionary biology}.
\newblock Cambridge, UK: Cambridge University Press.

\bibitem[Corthay, 2014]{Corthay2014}
Corthay, A. (2014).
\newblock Does the immune system naturally protect against cancer?
\newblock {\em Frontiers in immunology}, 5:1--8.

\bibitem[Dawkins, 1976]{Dawkins1976}
Dawkins, R. (1976).
\newblock {\em The selfish gene}.
\newblock Oxford, UK: Oxford University Press.

\bibitem[Deacon, 1998]{Deacon1998}
Deacon, T.~W. (1998).
\newblock {\em The symbolic species: The co-evolution of language and the
  brain}.
\newblock New York, NY: W. W. Norton.

\bibitem[Dennett, 1995]{Dennett1995}
Dennett, D.~C. (1995).
\newblock {\em Darwin's dangerous idea: Evolution and the meanings of life}.
\newblock New York, NY: Simon {\&} Schuster.

\bibitem[Diamond, 1997]{Diamond1997}
Diamond, J. (1997).
\newblock {\em Guns, germs, and steel: The fates of human societies}.
\newblock New York, NY: W. W. Norton.

\bibitem[Forterre, 2005]{Forterre2005}
Forterre, P. (2005).
\newblock The two ages of the {RNA} world, and the transition to the {DNA}
  world: A story of viruses and cells.
\newblock {\em Biochimie}, 87:793--803.

\bibitem[Godfrey-Smith, 2007]{Godfrey-Smith2007}
Godfrey-Smith, P. (2007).
\newblock Conditions for evolution by natural selection.
\newblock {\em The Journal of Philosophy}, 104:489--516.

\bibitem[Godfrey-Smith, 2011]{Godfrey-Smith2011}
Godfrey-Smith, P. (2011).
\newblock Darwinian populations and transitions in individuality.
\newblock In Calcott, B. and Sterelny, K., editors, {\em The major transitions
  in evolution revisited}, pages 65--81. Cambridge, MA: MIT Press.

\bibitem[Hamilton, 1964]{Hamilton1964}
Hamilton, W.~D. (1964).
\newblock The genetical evolution of social behaviour: {I} and {II}.
\newblock {\em Journal of Theoretical Biology}, 7:1--16 and 17--52.

\bibitem[Hammerstein, 2003]{Hammerstein2003}
Hammerstein, P. (2003).
\newblock {\em Genetic and cultural evolution of cooperation}.
\newblock Cambridge, MA: MIT Press.

\bibitem[Harari, 2015]{Harari2015}
Harari, Y.~N. (2015).
\newblock {\em Sapiens: A brief history of humankind}.
\newblock New York, NY: HarperCollins.

\bibitem[Hilbert, 2012]{Hilbert2012}
Hilbert, M. (2012).
\newblock How much information is there in the ``information society''?
\newblock {\em Significance}, 4:8--12.

\bibitem[Hilbert and L\'opez, 2011]{Hilbert2011}
Hilbert, M. and L\'opez, P. (2011).
\newblock The world's technological capacity to store, communicate, and compute
  information.
\newblock {\em Science}, 6025:60--65.

\bibitem[Hinton and Nowlan, 1987]{Hinton1987}
Hinton, G.~E. and Nowlan, S.~J. (1987).
\newblock How learning can guide evolution.
\newblock {\em Complex Systems}, 1:495--502.

\bibitem[Keightley and Otto, 2006]{Keightley2006}
Keightley, P.~D. and Otto, S.~P. (2006).
\newblock Interference among deleterious mutations favours sex and
  recombination in finite populations.
\newblock {\em Nature}, 443:89--92.

\bibitem[Koestler, 1967]{Koestler1967}
Koestler, A. (1967).
\newblock {\em The ghost in the machine}.
\newblock London, UK: Hutchinson.

\bibitem[Margulis, 1970]{Margulis1970}
Margulis, L. (1970).
\newblock {\em Origin of eukaryotic cells}.
\newblock New Haven, CT: Yale University Press.

\bibitem[Margulis, 1981]{Margulis1981}
Margulis, L. (1981).
\newblock {\em Symbiosis in cell evolution}.
\newblock San Francisco, CA: W. H. Freeman.

\bibitem[Maynard~Smith and Szathm\'ary, 1995]{MaynardSmith1995}
Maynard~Smith, J. and Szathm\'ary, E. (1995).
\newblock {\em The major transitions in evolution}.
\newblock Oxford, UK: Oxford University Press.

\bibitem[McShea and Brandon, 2011]{McShea2011}
McShea, D.~W. and Brandon, R.~N. (2011).
\newblock {\em Biology's first law: The tendency for diversity and complexity
  to increase in evolutionary systems}.
\newblock Chicago, IL: University of Chicago Press.

\bibitem[Nolan and Lenski, 2010]{Nolan2010}
Nolan, P. and Lenski, G.~E. (2010).
\newblock {\em Human societies: An introduction to macrosociology}.
\newblock Boulder, CO: Paradigm Publishers, 11th edition.

\bibitem[Richerson and Boyd, 2005]{Richerson2005}
Richerson, P.~J. and Boyd, R. (2005).
\newblock {\em Not by genes alone: How culture transformed human evolution}.
\newblock Chicago, IL: University of Chicago Press.

\bibitem[Simon, 1962]{Simon1962}
Simon, H.~A. (1962).
\newblock The architecture of complexity.
\newblock {\em Proceedings of the American Philosophical Society},
  106:467--482.

\bibitem[Soros and Stanley, 2014]{Soros2014}
Soros, L.~B. and Stanley, K.~O. (2014).
\newblock Identifying necessary conditions for open-ended evolution through the
  artificial life world of chromaria.
\newblock In Sayama, H., Rieffel, J., Risi, S., Doursat, R., and Lipson, H.,
  editors, {\em 14th International Conference on the Synthesis and Simulation
  of Living Systems}.

\bibitem[Sterelny, 2011]{Sterelny2011}
Sterelny, K. (2011).
\newblock Evolvability reconsidered.
\newblock In Calcott, B. and Sterelny, K., editors, {\em The major transitions
  in evolution revisited}, pages 83--100. Cambridge, MA: MIT Press.

\bibitem[Szathm\'ary, 2015]{Szathmary2015}
Szathm\'ary, E. (2015).
\newblock Toward major evolutionary transitions theory 2.0.
\newblock {\em Proceedings of the National Academy of Sciences},
  112:10104--10111.

\bibitem[Taylor, 2015]{Taylor2015}
Taylor, T. (2015).
\newblock Requirements for open-ended evolution in natural and artificial
  systems.
\newblock In Beslon, G., Elena, S., Hogeweg, P., Schneider, D., and Stepney,
  S., editors, {\em EvoEvo Workshop at the 13th European Conference on
  Artificial Life}.

\bibitem[Taylor et~al., 2016]{Taylor2016}
Taylor, T., Bedau, M., Channon, A., Ackley, D., et~al. (2016).
\newblock Open-ended evolution: Perspectives from the {OEE} workshop in {York}.
\newblock {\em Artificial Life}, 22:408--423.

\bibitem[Turney, 1989]{Turney1989}
Turney, P.~D. (1989).
\newblock The architecture of complexity: A new blueprint.
\newblock {\em Synthese}, 79:515--542.

\end{thebibliography}

\end{document}